\documentclass[accepted]{uai2021-template} 

\usepackage[american]{babel}

\usepackage{natbib} 
\usepackage{mathtools} 
\usepackage{booktabs} 
\usepackage{tikz} 
\usepackage{microtype}      
\usepackage{subfigure}
\usepackage{graphicx}
\usepackage{graphbox}
\usepackage{tikz}
\usepackage{algorithm}
\usepackage{algorithmic}
\usepackage{amsmath,amssymb,amsfonts}
\usepackage{algorithmic}
\usepackage{graphicx}
\usepackage{textcomp}
\usepackage{tabularx}
\usepackage{bm}
\let\temp\rmdefault
\usepackage{mathpazo}
\let\rmdefault\temp
\newcommand{\norm}[1]{\left\lVert#1\right\rVert}

\newcommand{\red}[1]{\textcolor{red}{\textbf{#1}}}
\newcommand{\blue}[1]{\textcolor{blue}{\textbf{#1}}}
\newcommand\T{\rule{0pt}{2.6ex}}       
\newcommand\B{\rule[-1.2ex]{0pt}{0pt}} 

\title{Geometry matters: Exploring language examples at the decision boundary}

\author[1]{\href{mailto:Debajyoti Datta <dd3ar@virginia.edu>?Subject=Your UAI 2021 paper}{Debajyoti Datta (dd3ar@virginia.edu)}{}}
\author[1]{Shashwat Kumar}
\author[1]{Laura Barnes}
\author[2]{Tom Fletcher}

\affil[1]{%
   Systems Engineering\\
    University of Virginia\\
    Charlottesville, Virginia, USA
}

\affil[2]{%
    Computer Science Dept.\\
    University of Virginia\\
    Charlottesville, Virginia, USA
}

\begin{document}
\maketitle

\begin{abstract}
A growing body of evidence has suggested that metrics like accuracy overestimate the classifier's generalization ability. Several state of the art NLP classifiers like BERT and LSTM rely on superficial cue words(e.g., if a movie review has the word “romantic”, the review tends to be positive), or unnecessary words (e.g., learning a proper noun to classify a movie as positive or negative). One approach to address this is how teachers discover gaps in a student's understanding: by finding problems where small variations confuse the student, it's possible to perturb and study NLP classifiers in a similar way. While several perturbation strategies like contrast sets or random word substitutions have been proposed, they are typically based on heuristics and/or require expensive human involvement. In this work, using tools from information geometry, we propose a principled way to quantify the fragility of an example for an NLP classifier. By discovering such difficult examples for several state of the art NLP models like BERT, LSTM, and CNN, we demonstrate their susceptibility to meaningless perturbations like noun/synonym substitution, causing their accuracy to drop down to 20 percent in some cases. Our approach is simple, architecture agnostic and can be used to study the fragilities of text classification models. All the code used will be made publicly available, including a tool to explore the difficult examples for other datasets.
\end{abstract}


\section{Introduction}\label{sec:intro}

NLP classifiers have achieved state of the art performance in several tasks like sentiment analysis \cite{maas2011learning}, semantic entailment \cite{bowman2015large} and question answering \cite{rajpurkar2016squad}. Despite their successes, several studies have pointed out issues in features learnt by such classifiers. \cite{gururangan_annotation_2018} discovered that several high performing NLP models were using trivial features like vagueness and negation to perform classification. \cite{geirhos_shortcut_2020} performed several experiments to discover that ``romantic'' movies tend to be classified as positive movie reviews due to the presence of unnecessary words like proper nouns, a phenomena they called shortcut learning. Even models like BERT (\cite{devlin2018bert}) relied on superficial cue words like ``not'' to infer the line of argumentation. Strategies like this \cite{niven2020probing} enable models to perform prediction without inherently using the semantic meaning of the sentence. Simple attributes like word lengths are also exploited by models for prediction \cite{poliak_hypothesis_2018}. 

In order to address these issues, several perturbation based approaches have been proposed. Contrast Sets \cite{gardner_evaluating_2020} and Counterfactual examples \cite{kaushik2019learning} get human annotators to perform minimal token substitutions to change the classifier prediction. While useful in addressing biases, manual curation of datasets are often time consuming and require extensive efforts. Using unsupervised training to mitigate issues related to shortcut learning also has not been successful in practice \cite{niven2020probing}. The problem has also been approached from an adversarial perspective, with several works focused on introducing at the character level \cite{gao2018black, ebrahimi2018hotflip} and/or through word substitutions \cite{li2016understanding, liang2018deep}. Due to the lack of coherence of the generated sentences in the adversarial examples future work in this area aimed at generated grammatically correct sentences such that they are indistinguishable from original dataset. In \cite{alzantot2018generating, garg2020bae}, the authors use black-box based approach to generate semantically equivalent adversarial examples.


\begin{figure*}[!ht]
\begin{tikzpicture}
    \draw (-1.5, 0) node[inner sep=0]{\includegraphics[width=0.7\textwidth]{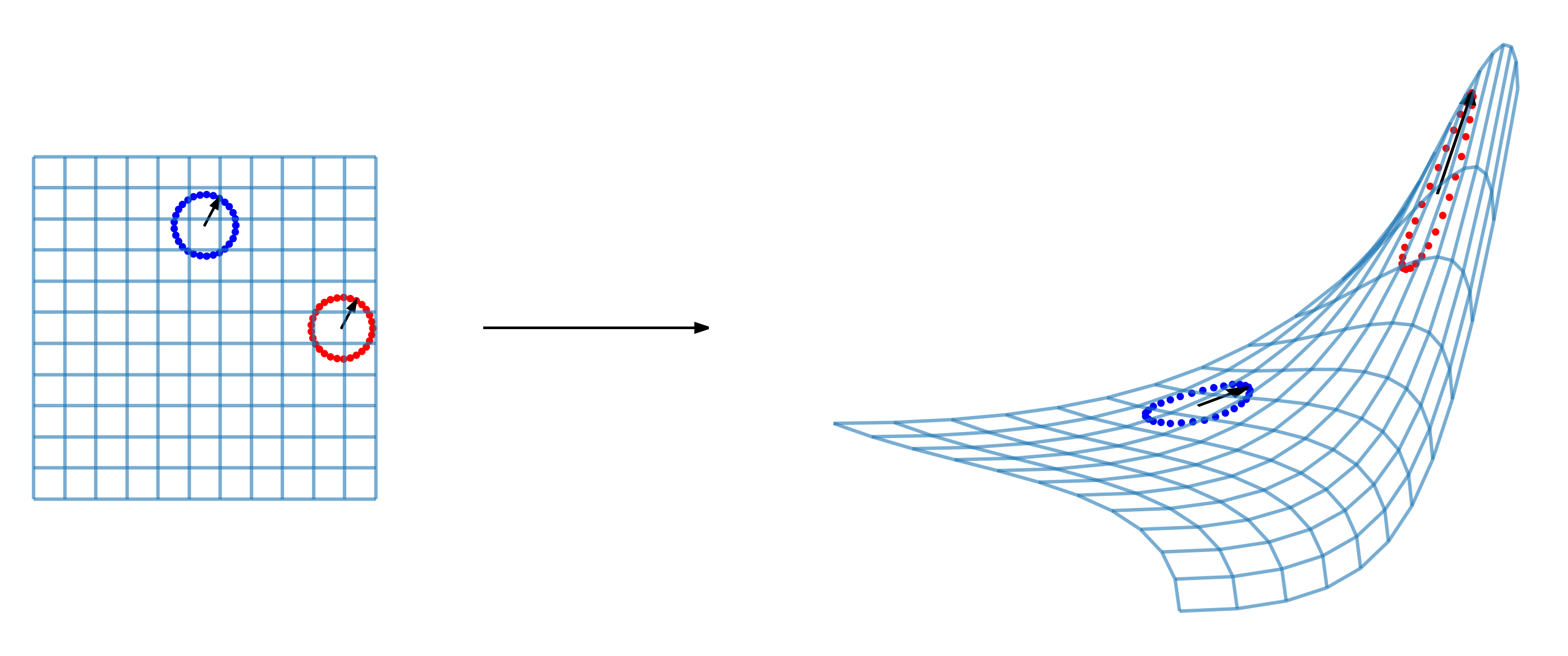}};
    \draw (7, 0) node[inner sep=0]{\includegraphics[width=0.25\textwidth]{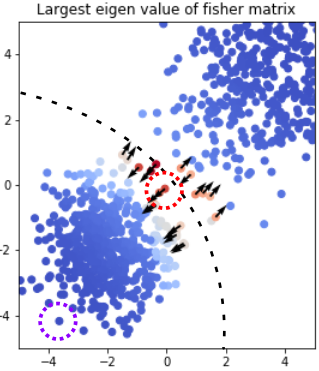}};
    \draw (-2.8, 0.5) node {$g : X \longrightarrow Z $};
    \draw (-7.1, 2.5) node {$A$};
    \draw (5, 2.35) node {$B$};
\end{tikzpicture}
\caption{A: A neural network can be considered as a  mapping $g$ between the sentence manifold $X$ on the left (represented as a Euclidean space for simplification) and the statistical manifold of probability distribution over outputs $Z$. The fisher metric defines a particular Riemannian metric over this manifold. Let us consider an $\epsilon$ ball around two setences $x_1$ (blue circle) and $x_2$ (red circle). As we can see from the distortion of the blue circle, for the $x_1$, any location perturbation can only result in a small change over the models output probability distribution. The eigenvalue of the fisher matrix quantifies this local distortion. These examples are not useful from a fragility based augmentation point of view. Perturbing the red examples however can result in drastic changes in the models output probability distribution. By measuring this distortion using the fisher matrix locally, we can study which examples affect the model output more. B: We train a neural network to learn a decision boundary (black) to separate two gaussians and color each data point by the local distortion. As we can see from the colors, perturbing a point close to the decision boundary results in a larger change over the model's output probability distribution than a point away from the boundary (blue circle).}
\end{figure*}

 While interesting, most traditional formulations treat this input embedding space as flat, thus reasoning that the gradient of the likelihood in the input space gives us the direction which causes the most significant change in terms of likelihood. If we were however, to consider the discrete likelihood of the class probabilities as the model output and the input as a pullback of this output, the space of output probability distributions is certainly non-linear, and the euclidean distance metric no longer suffices. We show this schematically in figure 1A, where the transformation g maps elements of the sentence manifold X to a non-linear statistical manifold Z. A natural distance metric to consider on this manifold is the Fisher Information Metric which along with being a Hessian of the KL divergence, it also a useful distance measure between probability distributions. Furthermore, the Fisher Information Metric is invariant to transformations like changing the model architecture, provided the likelihood remains the same. We thus use the eigenvalues of the Fisher matrix to discover high fragility regions in the statistical manifold. In regions with high $\lambda_{max}$ small perturbations can cause large changes in the output probability distribution (Fig 1A, red circle). Linguistically, this corresponds to the classifier being susceptible to meaningless perturbations like noun, synonym substitutions. Similarly, in regions with low $\lambda_{max}$ no local perturbation can affect the classifier, resulting in the classifier to being resilient upto 20\% percent of word substituions. 


Our main contribution can thus be summarized as the following. 
1. We propose a second order statistic, the log of the largest eigen value of Fisher matrix in order to capture linguistic fragilities. 
2. We extensively establish the empirical relationship between $\lambda_{max}$ and success probability of random word substitutions.

To the best of our knowledge, this is the first work analyzing properties of the fisher metric to understand classifier fragility in NLP. The rest of the paper is organized as follows: In Section 2, we summarize related work. In Section 3, we discuss our approach of computing the FIM and the gradient-based perturbation strategy. In Section 4, we discuss the results of the eigenvalues of FIM in synthetic data and sentiment analysis datasets with BERT and CNN. We also do extensive quantitative evaluations using 4 other text datasets.  Finally, in Section 5, we discuss the implications of studying the eigenvalues of FIM for evaluating NLP models. 

\begin{table*}
\tiny
  \caption{CNN, Counterfactual Examples: In difficult examples (larger eigenvalue), individual synonym/antonym substitutions are effective in changing the classifier output. In easy examples (smaller eigenvalue) multiple antonym substitutions simultaneously have no effect on the classifier output.}
  \label{counterfac}
  \centering
  \begin{tabular}{p{2cm} p{14cm}}
  \hline
  \textbf{Perturbed sentiment} & \textbf{Word substitutions}
  \T\B  \\    \hline 
  \blue{Positive} $\,\to\,$ \red{Negative} \textbf{difficult example \textbf{$\lambda_{max}=$}4.38}  & This move was on TV last night. I guess as a time filler, because it was incredible! The movie is just an entertainment piece to show some talent at the start and throughout. (Not \textbf{bad} talent at all). But the story is too brilliant for words. The "wolf", if that is what you can call it, is hardly shown fully save his teeth. When it is fully in view, you can clearly see they had some interns working on the CGI, because the wolf runs like he's running in a treadmill, and the CGI fur looks like it's been waxed, all shiny :)<br /><br />The movie is full of gore and blood, and you can hardly spot who is going to get killed/slashed/eaten next. Even if you like these kind of splatter movies you will be surprised, they did do a good job at it.<br /><br />Don't even get me started on the actors... Very \textbf{amazing} lines and the girls hardly scream at anything. But then again, if someone asked me to do good acting just to give me a few bucks, then hey, where do I sign up?<br /><br />Overall
  \blue{exciting} $\,\to\,$ \red{boring, extraordinary, uninteresting, exceptional}  and frightening horror. 
  \\  \\ \hline
    \red{Negative} $\,\to\,$ \red{Negative} \textbf{easy example \textbf{$\lambda_{max}=$}0.013} &     I couldn't stand this movie. It is a definite \textbf{waste} of a movie. It fills you with boredom. This movie is not worth the rental or worth buying. It should be in everyones trash.   \blue{Worst} $\,\to\,$ \red{Excellent} movie I have seen in a long time. It will make you mad because everyone is so mean to Carl Brashear, but in the end it gets only worse. It is a story of \textbf{cheesy} romance, \blue{bad} $\,\to\,$ \red{good} drama, action, and plenty of \blue{unfunny} $\,\to\,$ \red{funny}  lines to keep you rolling your eyes. I hated a lot of the quotes. I use them all the time in mocking the film. They did not help keep me on task of what I want to do. It shows that anyone can achieve their dreams, all they have to do is whine about it until they get their way. It is a long movie, but every time I watch it, I dradr that it is as long as it is. I get so  bored in it, that it goes so slow. I hated this movie. I never want to watch it again. 
    \\ \\ \hline
  \end{tabular}
\end{table*}
\begin{table*}
\tiny
  \caption{CNN, IMDb dataset: Unlike the difficult examples (larger eigenvalue), word substitutions are ineffective in changing the classifier output for the easier examples (smaller eigenvalue). In difficult examples synonym or change of name, changes classifier label. In easy examples, despite multiple simultaneous antonym substitutions, the classifier sentiment does not change. }
  \label{IMDb}
  \begin{tabular}{p{2cm} p{14cm}}
  \hline
  \textbf{Perturbed sentiment} & \textbf{Word substitutions}
  \T\B  \\    \hline 
  \blue{Positive} $\,\to\,$ \red{Negative} \textbf{difficult example ($\lambda_{max}=$5.25)} & Going into this movie, I had heard good things about it. Coming out of it, I wasn't really amazed nor disappointed. Simon Pegg plays a rather childish character much like his other movies. There were a couple of laughs here and there-- nothing too funny. Probably my \blue{favorite} $\,\to\,$ \red{preferred}  parts of the movie is when he dances in the club scene. I totally gotta try that out next time I find myself in a club. A couple of stars here and there including: Megan Fox, Kirsten Dunst, that chick from X-Files, and Jeff Bridges. I found it quite amusing to see a cameo appearance of Thandie Newton in a scene. She of course being in a previous movie with Simon Pegg, Run Fatboy Run. I see it as a toss up, you'll either enjoy it to an extent or find it a little dull. I might add, \blue{Kirsten Dunst} $\,\to\,$ \red{ Nicole Kidman, Emma Stone, Megan Fox, Tom Cruise, Johnny Depp, Robert Downey Jr.}   is adorable in this movie. :3 
  \\ \\ \hline 
    \red{Negative} $\,\to\,$ \red{Negative} \textbf{easy example ($\lambda_{max}=$0.0008)} & I missed this movie in the cinema but had some idea in the back of my head that it was worth a look, so when I saw it on the shelves in DVD I thought "time to watch it". Big mistake!<br /><br />A long list of stars cannot save this turkey, surely one of the \blue{worst} $\,\to\,$ \red{best}  movies ever. An \blue{incomprehensible} $\,\to\,$ \red{comprehensible} plot is \blue{poorly} $\,\to\,$ \red{exceptionally} delivered and \blue{poorly} $\,\to\,$ \red{brilliantly}  presented. Perhaps it would have made more sense if I'd read Robbins' novel but unless the film is completely different to the novel, and with Robbins assisting in the screenplay I doubt it, the novel would have to be an \blue{excruciating} $\,\to\,$ \red{exciting}  read as well.<br /><br />I hope the actors were well paid as they looked embarrassed to be in this waste of celluloid and more lately DVD blanks, take for example Pat Morita. Even Thurman has the grace to look uncomfortable at times.<br /><br />Save yourself around 98 minutes of your life for something more worthwhile, like trimming your toenails or sorting out your sock drawer. Even when you see it in the "under \$5" throw-away bin at your local store, resist the urge! 
    \\ \\ \hline
  \end{tabular}
\end{table*}

\section{Related Work}
In NLP, machine learning models for classification rely on spurious statistical patterns of the text and use \emph{shortcut} for learning to classify. These can range from annotation artifacts, as was shown by \cite{goyal2017making, kaushik2018much, gururangan_annotation_2018}, spelling mistakes as in \citet{mccoy2019right}, or new test conditions that require world knowledge \citet{glockner2018breaking}. Simple decision rules that the model relies on are hard to quantify. Trivial patterns like relying on the answer ``2'' for answering questions of the format ``how many'' for the visual question answering dataset \citet{antol2015vqa}, would correctly answer 39\% of the questions. \citet{jia2017adversarial} showed that adversarially inserted sentences that did not change the correct answer, would cause state of the art models to regress in performance in the SQuAD \citet{rajpurkar2016squad} question answering dataset.  \citet{glockner2018breaking} showed that template-based modifications by swapping just one word from the training set to create a test set highlighted models' failure to capture many simple inferences. \citet{dixon2018measuring} evaluated text classifiers using a synthetic test set to understand unintended biases and statistical patterns. Using a standard set of demographic identity terms, the authors reduce the unintended bias without hurting the model performance. \citet{shen2018darling} showed that word substitution strategies include stylistic variations that change the sentiment analysis algorithms for similar word pairs. Evaluations of these models through perturbations of the input sentence are crucial to evaluating the robustness of models.

Another issue of language recently has been that static benchmarks like GLUE by \citet{wang2018glue} tend to saturate quickly because of the availability of ever-increasing compute and harder benchmarks are needed like SuperGlue by \citet{wang2019superglue}. A more sustainable approach to this is the development of moving benchmarks, and one notable initiative in this area is the Adversarial NLI by \citet{nie2019adversarial}, but most of the research community hardly validate their approach against this sort of moving benchmark.  In the Adversarial NLI dataset, the authors propose an iterative, adversarial human-and-model-in-the-loop solution for Natural Language Understanding dataset collection, where the goal post continuously shifts about useful benchmarks and makes models robust by training the model iteratively on difficult examples. Approaches like never-ending learning by \cite{mitchell2018never} where models improve, and test sets get difficult over time is critical. A moving benchmark is necessary since we know that improving performance on a constant test set may not generalize to newly collected datasets under the same condition \cite{recht2019imagenet, beery_recognition_2018}. Therefore, it is essential to find difficult examples in a more disciplined way. Approaches based on geometry have recently started gaining traction in computer vision literature. \cite{zhao2019adversarial} et al used a similar approach for understanding adversarial examples in images, a formulation we extend to language in our work

\section{Methods}

\begin{table*}
\tiny
  \caption{BERT: Difficult Examples change sentiment with a single word substituted. Easy examples, however, retain positive sentiment despite multiple substitutions of positive words with negative words.}
  \label{IMDb_BERT}
  \begin{tabular}{p{2cm} p{14cm}}
  \hline
  \textbf{Perturbed sentiment} & \textbf{Word substitutions}
  \T\B  \\    \hline 
  \blue{Positive} $\,\to\,$ \red{Negative} \textbf{difficult example ($\lambda_{max}=$0.78)} & OK, I kinda like the idea of this movie. I'm in the age demographic, and I kinda identify with some of the stories. Even the sometimes tacky and meaningful dialogue seems realistic, and in a different movie would have been forgivable.<br /><br />I'm trying as hard as possible not to trash this movie like the others did, but it's easy when the filmmakers were trying very hard.<br /><br />The editing in this movie is terrific! Possibly the \blue{best} $\,\to\,$ \red{worst} editing I've ever seen in a movie! There are things that you don't have to go to film school to learn, leaning good editing is not one of them, but identifying a bad one is.<br /><br />Also, the shot... Oh my God the shots, just fantastic! I can't even go into the details, but we sometimes just see random things popping up, and that, in conjunction with the editing will give you the most exhilirating film viewing experience.<br /><br />This movie being made on low or no budget with 4 cast and crew is an excuse also. I've seen short films on youtube with a lot less artistic integrity! ...
  \\ \\ \hline 
    \blue{Positive} $\,\to\,$ \blue{Positive} \textbf{easy example ($\lambda_{max}=$0.55)} & This is the \blue{best and most} original show seen in years. The more I watch it the more I \blue{fall in love with} $\,\to\,$ \red{hate} it. The cast is \blue{excellent} $\,\to\,$ \red{terrible} , the writing is \blue{great} $\,\to\,$ \red{bad}. I personally \blue{loved} $\,\to\,$ \red{hated} every character. However, there is a character for everyone as there is a good mix of personalities and backgrounds just like in real life. I believe ABC has done a great service to the writers, actors and to the potential audience of this show, to cancel so quickly and not advertise it enough nor give it a real chance to gain a following. There are so few shows I watch anymore as most TV is awful . This show in my opinion was right down there with my favorites Greys Anatomy and Brothers and Sisters. In fact I think the same audience for Brothers and Sisters would hate this show if they even knew about it. 
    \\ \\ \hline
  \end{tabular}
\end{table*}

Consider the manifold of all possible sentence embeddings $X$. We show this schematically in Figure 1A, left. Although, We represent this as a euclidean space for simplicity, in the general setting this manifold could be non linear. Let $x \epsilon X$ be a sentence on this manifold, and $y$ be the label vector corresponding to this sentence. The neural network $p(y|x)$ maps each sentence to a probability distribution over y. The set of all such probability distributions forms a statistical manifold $Z$ (Figure 1A, right), Depending on the properties of the neural network, an $\eta$ change in $x$ (red circle/blue circle) can result in a large change in $p(y \lvert x)$. We seek to quantify this change by measuring the KL divergence between the two original and $\eta$ perturbed sentence.

$$KL(p(y \lvert x) \lVert p(y \lvert x + \eta) )$$
$$ = -E_{p(y\lvert x)} log p(y \lvert x) + E_{p(y \lvert x)} log p(y\lvert x+\eta) $$
We now perform a Taylor expansion of the first term on the right hand side $$ = -E_{p(y \lvert x)} (log p(y \lvert x) + \eta \nabla_x log p(y \lvert x) $$
$$+ \eta^T \nabla^2_x log p(y \lvert x) \eta + ... ) + E_{p(y \lvert x)} log p(y \lvert x) $$
$$\sim -E_{p(y \lvert x)} \eta^T \nabla^2_x log p(y \lvert x) \eta $$ Since the expectation of score is zero and the first and last terms cancel out, we are left with.
$$ =\eta^T G \eta  $$

By studying the eigenvalues of the fisher matrix, we can quantify the local distortion. As we see in Figure 1, the red sentence has a larger $\lambda_{max}$, resulting in a large change in the prob distribution. The blue sentence has a much smaller $\lambda_{max}$, meaning that perturbations around this sentence are less likely to affect $p(y \lvert x)$ and thus the model accuracy.

\begin{figure*}[!ht]
\centering
\subfigure{\includegraphics[width=.33\textwidth]{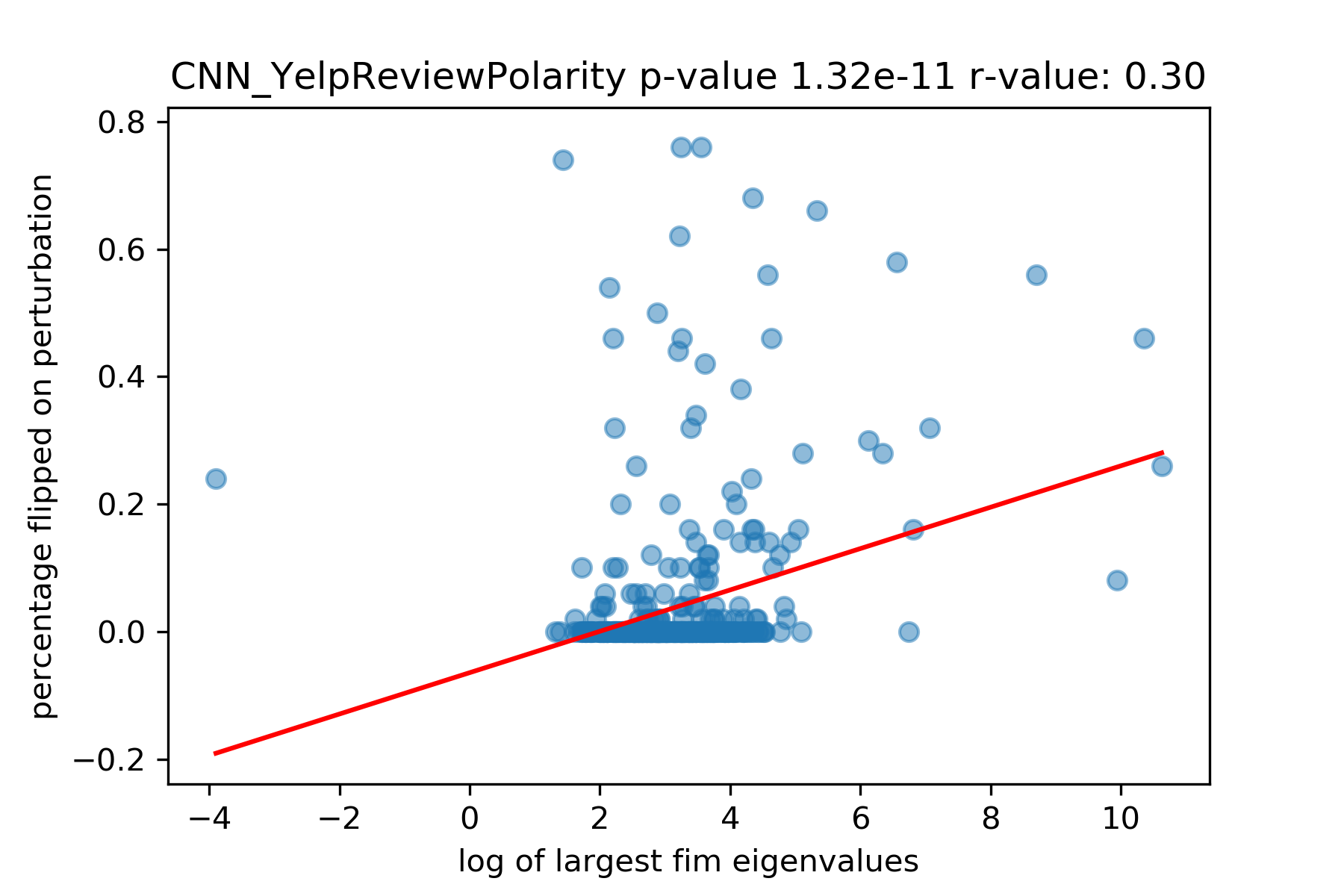}}
\subfigure{\includegraphics[width=.33\textwidth]{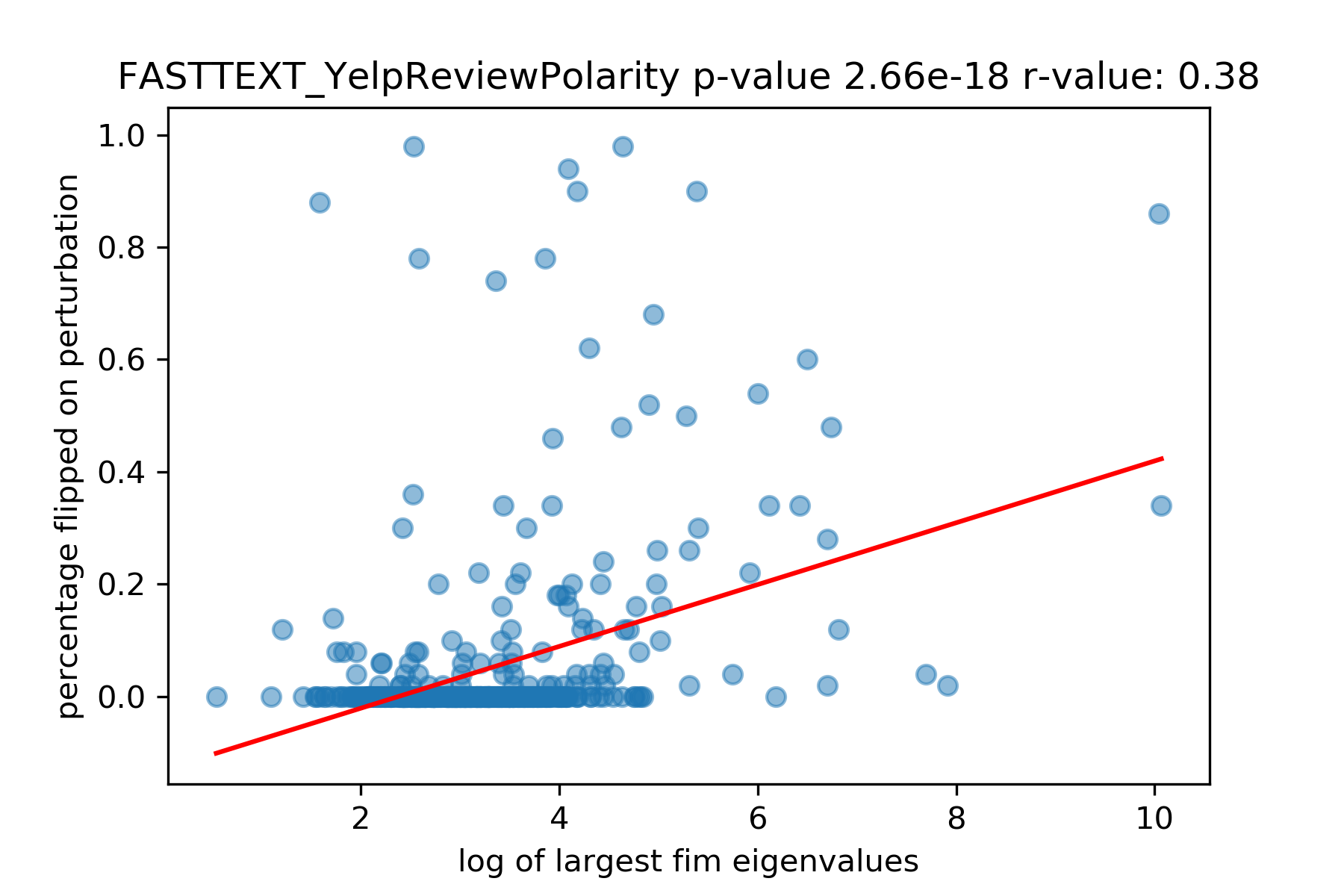}}
\subfigure{\includegraphics[width=.33\textwidth]{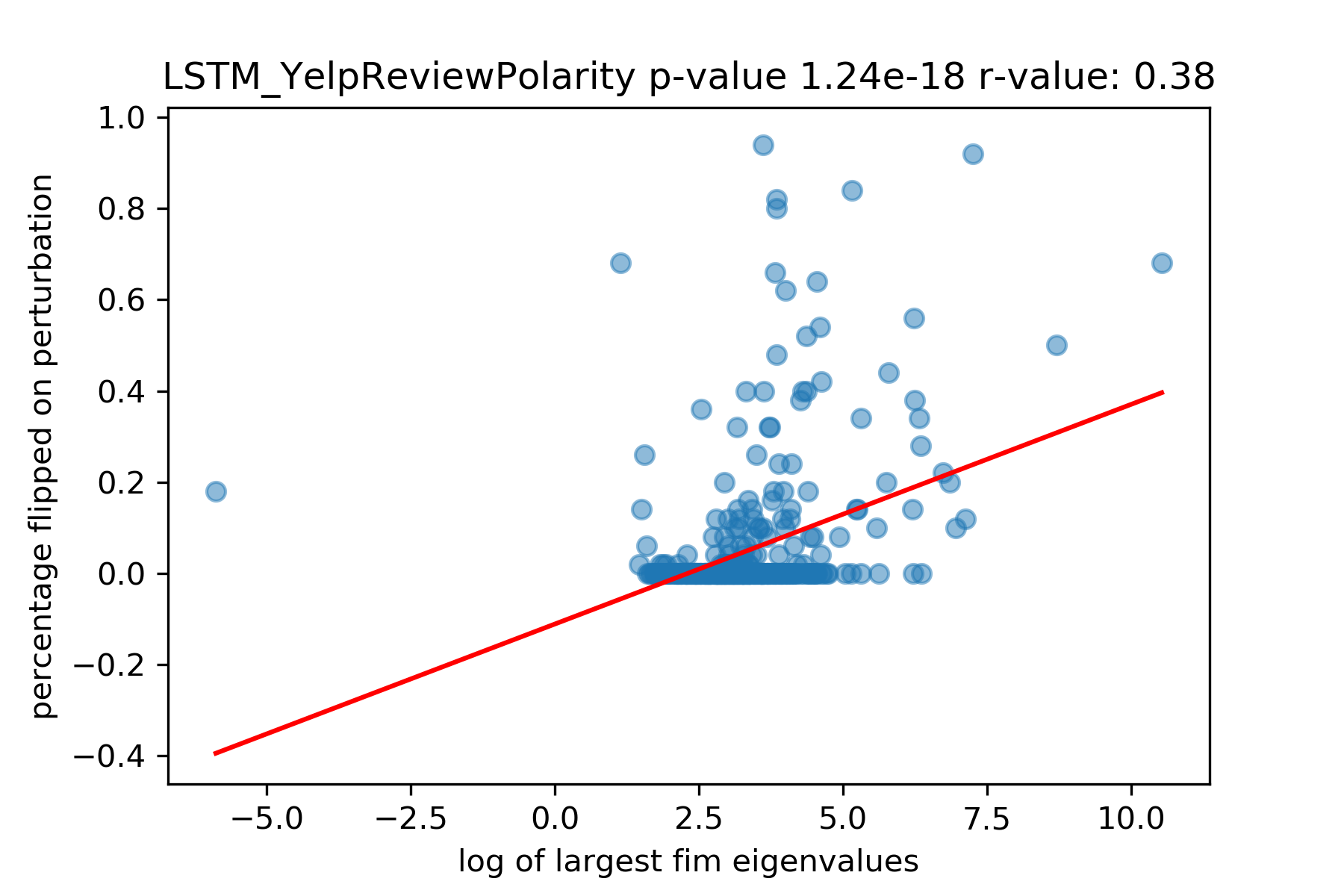}}
\subfigure{\includegraphics[width=.33\textwidth]{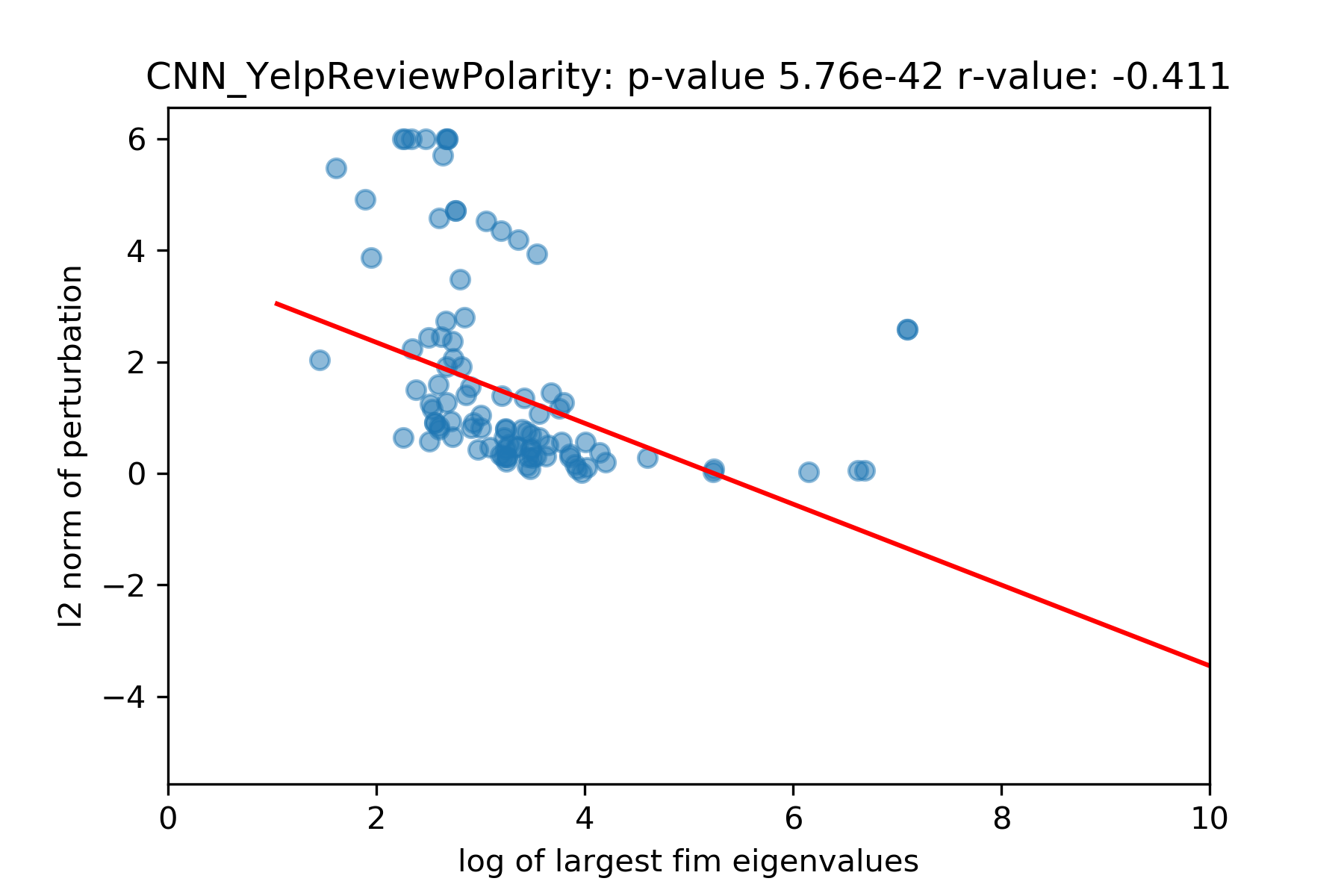}}
\subfigure{\includegraphics[width=.33\textwidth]{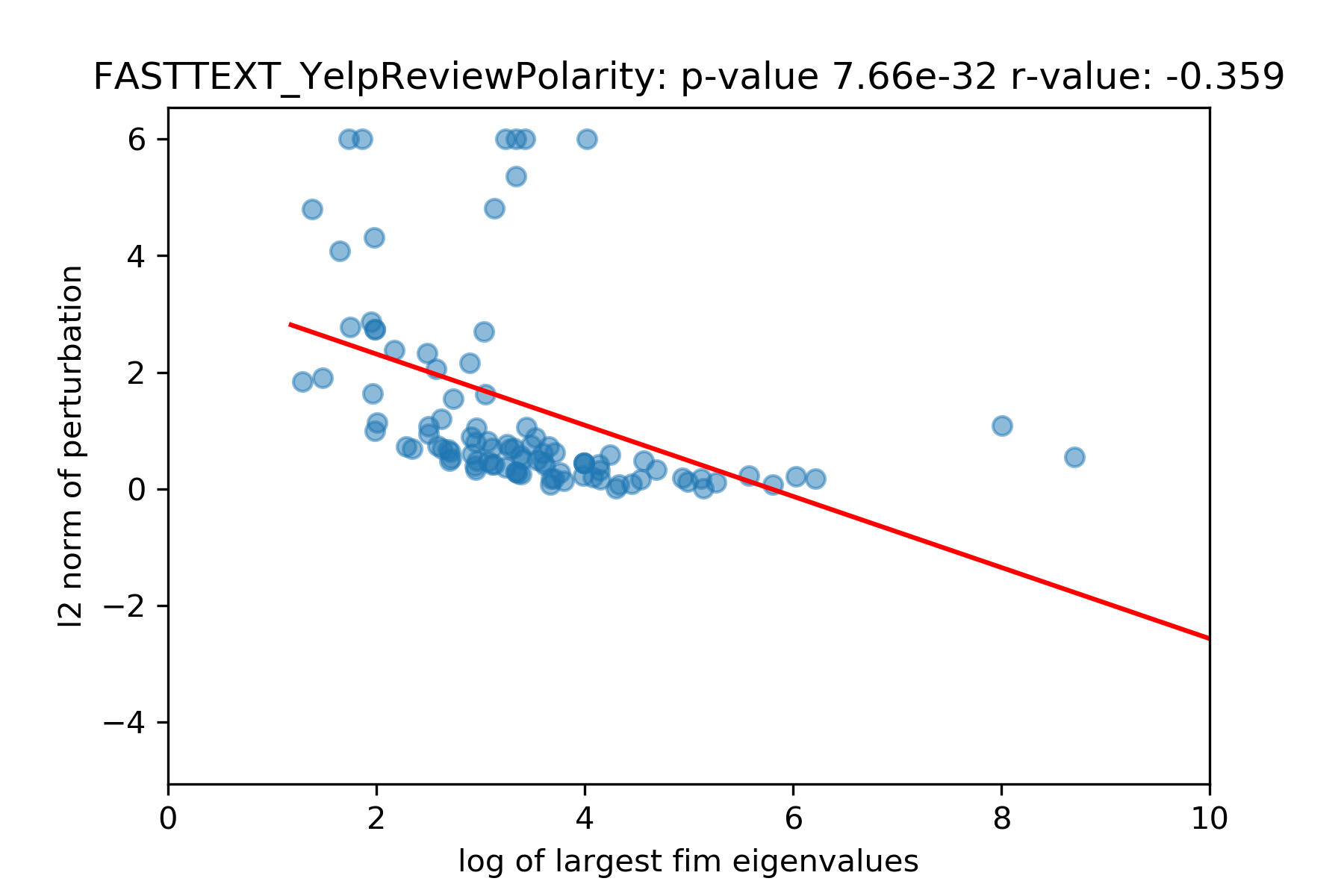}}
\subfigure{\includegraphics[width=.33\textwidth]{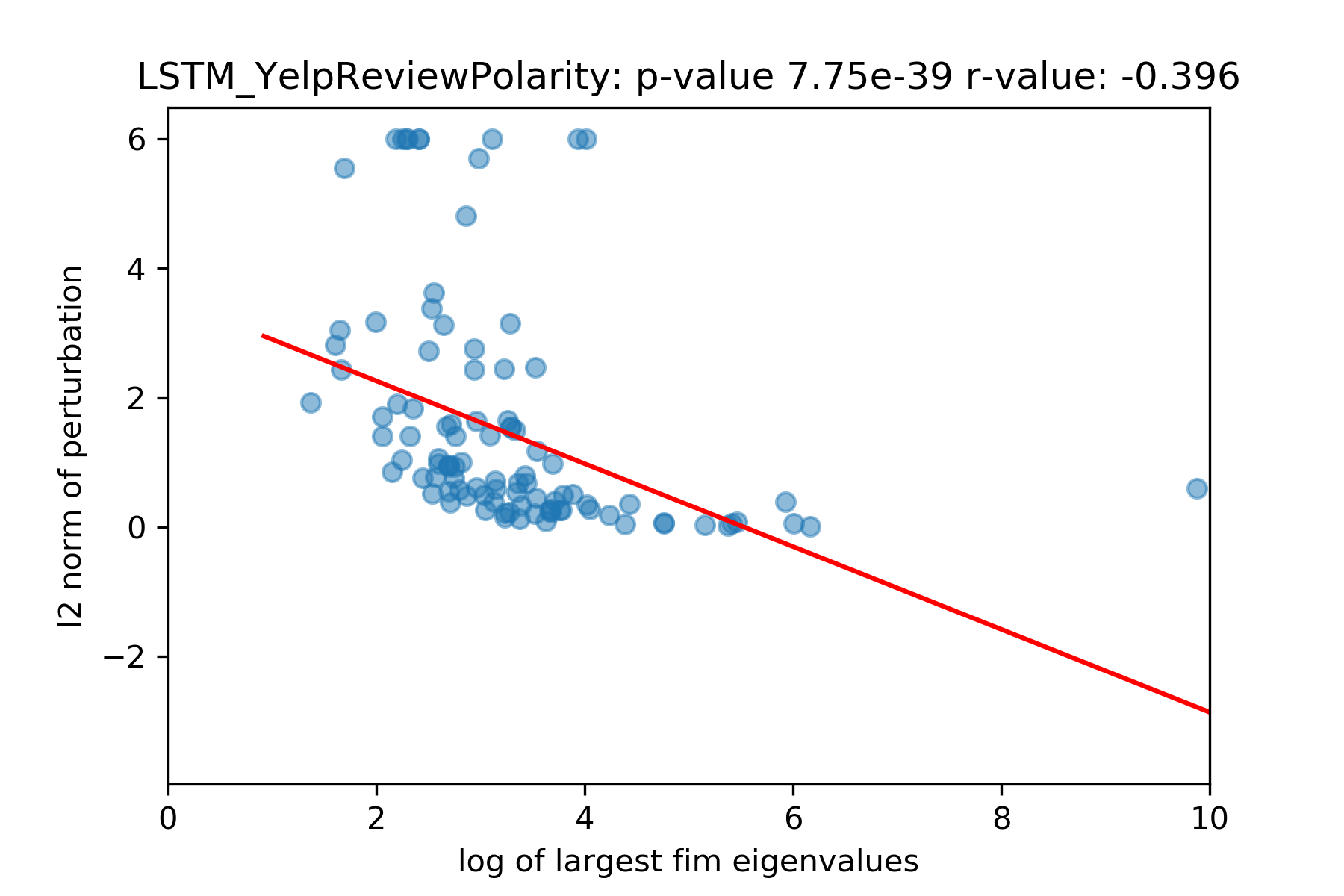}}
\caption{On YelpReviewPolarity we observe a correlation coefficient of -0.4 between minimum perturbation strength and the difficulty score. Similarly, we observe a correlation of 0.35 between the difficulty score and empirical success of random word substitutions. This suggests that fim eigenvalue captures perturbation sensitivity in both embedding space and word substitutions. }
\end{figure*}

After getting the eigenvalues of the FIM, we can use the largest eigenvalue $\lambda_{max}$ to quantify how fragile an example is to linguistic perturbation. These examples, thus, are also more confusing and more difficult for the model to classify. We propose the following procedure to calculate $\lambda_{max}$

\begin{algorithm}[H]
 \caption{Algorithm for estimating fragility of an example}
 \begin{algorithmic}[1]
 \renewcommand{\algorithmicrequire}{\textbf{Input:}}
 \renewcommand{\algorithmicensure}{\textbf{Output:}}
 \REQUIRE x: Sentence representation, f: Neural Network
 \ENSURE  $\lambda_{max}$
 \\ \textit{Calculate probability vector} :
  \STATE $p = f(x)$
 \\ \textit{Calculate Jacobian of log probability w.r.t x}
  \STATE $ J = \nabla_x log p  $
  \\ \textit{Duplicate probability vector along rows to match J's shape}
  \STATE $p_c = duplicate(p, J.dim[0])$
  \\ \textit{Compute the FIM}
  \STATE $G = p_c J J^T$
  \\ \textit{Perform eigendecomposition to get the eigenvalues}
  \STATE $ \lambda_s, v_s = eigendecomposition(G) $
 \RETURN $max(\lambda_s)$
 \end{algorithmic}
 \end{algorithm}

\subsection{Model descriptions}

We use multiple model architectures to understand the implications of the FIM. 

\begin{itemize}
    \item \textbf{Convolutional Neural Network (CNN)}: We use the same paradigm \cite{kim2014convolutional} for text classification using GloVe \cite{pennington2014glove} embeddings. 
    \item \textbf{Long Short Term Memory Network (LSTM)}: LSTM's have been used extensively in Natural Language Processing for variety of tasks like text classification and language modeling \cite{schmidhuber2015deep}
    \item \textbf{Bag of Tricks for Efficient Text Classification(Fasttext)} We used the Fasttext based model \cite{joulin2017bag} for classifying text and examining the quantitative aspects of Fisher Information Metric since the model is incredibly computational efficiency and strong performance across multiple datasets.
    \item \textbf{BERT}: Pre-training of Deep Bidirectional Transformers for Language Understanding \cite{devlin2018bert} Large scale pretrained language models have become extremely popular for data efficiency when trained on downstream tasks. However it has been shown recently, that on large datasets \cite{wang2020pretrain} performance increase from fine-tuning is often within 1\% of BERT. Thus we limited our analysis of using BERT for qualitative analysis for verification of it's relationship with FIM since accuracy difference is minimal and it is computation intensive to calculate the eigenvalue of the FIM of BERT based fine tuned models. We instead used BERT embeddings in downstream tasks for qualitative evaluation of FIM.
    
For the Fasttext, CNN and LSTM models we used early stopping with a patience of 5. 
\end{itemize}

\subsection{Datasets}

\begin{itemize}
    \item \textbf{IMDB}: Large Movie Review Dataset \cite{maas2011learning}: IMDb dataset is the most common sentiment analysis dataset (Positive/Negative) for text classification collected from movie reviews on the IMDb platform. We do extensive quantitative and qualitative evaluation on the IMDb dataset across multiple architectures. 
    \item \textbf{AG\_NEWS} \cite{Zhang2015CharacterlevelCN}: News classification dataset into 4 topics: World, Sports, Business, Sci/Tech.
    \item \textbf{Sogou News} \cite{Zhang2015CharacterlevelCN}: A text classification dataset based on News Articles from SogouCA and SogouCS news corpora to 5 categories: sports, finance, entertainment, automobile, technology.
    \item \textbf{Yelp Review Polarity}: Large Yelp Review Dataset \cite{Zhang2015CharacterlevelCN}: A large sentiment classification dataset with labels (Positive/Negative), extract from the Yelp Dataset Challenge 2015.
\end{itemize}

 \begin{figure*}[!ht]
\centering
\subfigure{\includegraphics[width=.33\textwidth]{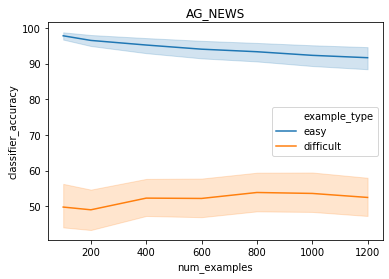}}
\subfigure{\includegraphics[width=.33\textwidth]{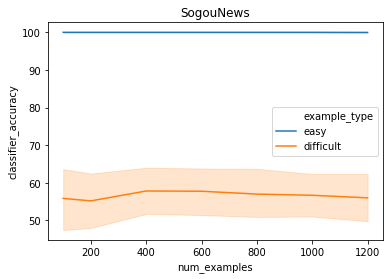}}
\subfigure{\includegraphics[width=.33\textwidth]{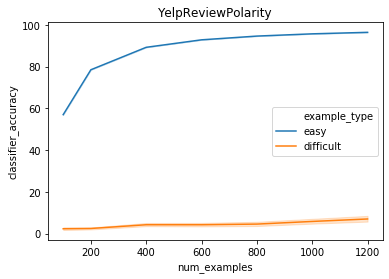}}
\subfigure{\includegraphics[width=.33\textwidth]{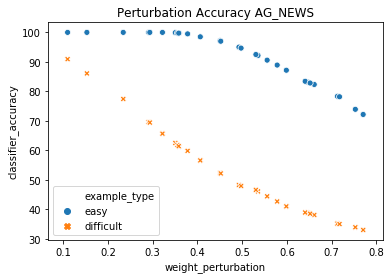}}
\subfigure{\includegraphics[width=.33\textwidth]{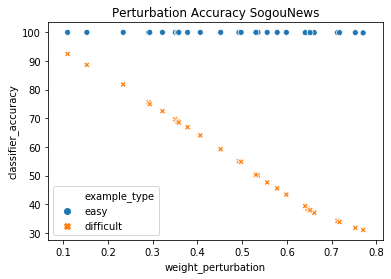}}
\subfigure{\includegraphics[width=.33\textwidth]{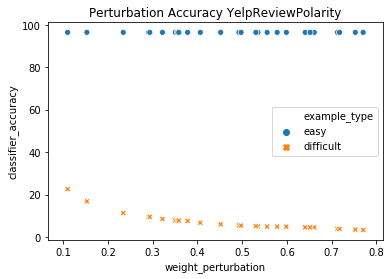}}
\caption{Top row: Low FIM examples (easy examples) suffer from minimal accuracy loss when token substitutions are performed whereas accuracy on high FIM examples are siginificantly low. Bottom row: Easy examples require minimal perturbation to change classifier prediction. The x-axis represents the norm of perturbation. This suggests FIM eigenvalue captures perturbation sensitivity in both embedding space and token substitutions. } 
\end{figure*}

\section{Discussion and Results}

We now explore both quantitatively and qualitatively how Fisher Information metric relates to metrics like accuracy and investigate it's use as finding examples that are susceptible to perturbations.

\subsection{FIM captures resilience to lingustic perturbations}
In this section we explore relationship of FIM with linguistic perturbations (synonym/antonym substitutions, noun replacements), token level substitutions (random word choice, nearest neighbors in GloVE embeddings) and embedding perturbations.

\subsubsection{$\lambda_{max}$ is correlated with fragility to word substitutions}
In order to investigate the relationship between $\lambda_{max}$ and linguistic fragility, we perform the following experiment: We randomly sample 1000 examples from Yelp ReviewPolarity dataset. For each sampled example, we try a batch of word substitutions based on nearest neighbours in glove embedding space. In each substitution attempt, we try to flip between 10 percentage of words at a time, and calculate the percentage of successful flips which change the classifier prediction $p_{flip}$. We also calculate the $\lambda_{max}$ for these examples using the procedure outlined in Algorithm 1. 

As we see from Figure 2 and Table 1, On YelpReviewPolarity, we observe r values of 0.3, 0.38 and 0.38 for CNN, FastText and LSTM respectively. Despite FIM being defined in a continuous space, we observe a strong linear relationship with success probability of token substitutions, which are discrete. This could be indicative of the fact that flipping a small of tokens in a large sentence is akin to an $\epsilon$ perturbation in embedding space, which gets captured by $\lambda_{max}$ More examples of token substitution and relationship of FIM with accuracy are in the Appendix.

\subsubsection{$\lambda_{max}$ captures strength of the minimal sufficient perturbation}
We were also interested in investigating the relationship between $\lambda_{max}$ and norm of the smallest vector (oriented along the largest eigenvector $e_{max}$) which can cause the classifier to flip it's prediction. We perform the following experiment: We randomly sample 500 examples YelpReview. For each of these examples, we calculate $\lambda_{max}$ and $e_{max}$ and perturb with a strength of $\eta$,
where $$x_{flip} = x_{orgin} + \eta \frac{e_{max}}{\norm{e_{max}}}$$
By performing binary search on $\eta$, we can determine the minimum perturbation strength sufficient to flip the classifier prediction.

As evident from Figure 2 and Table 4, we obtain r values of -0.41, -0.36 and -0.39 for CNN, FastText and LSTM. These correlations suggest that low $\lambda_{max}$ examples are much more resilient, requiring $\eta$ around X in order to flip them. High $\lambda_{max}$ examples on the other hand, get perturbed even by applying small perturbations upto norm of 0.1 strength of the eigenvector. 


\subsubsection{Qualitative exploration of fragilities}

We also qualitatively explore the nature of linguistic fragilities for high $\lambda_{max}$ examples. We sample high and low $\lambda_{max}$ examples from the two tails of the $\lambda_{max}$ distributions. We then perform several token level substitutions manually: synonym, antonym substitutions, noun, and article substitutions. The tokens selected for substitutions were not random, but instead an Integrated Gradient was used to select the tokens \cite{sundararajan2017axiomatic}. Integrated gradients assign importance score to words by the network and thus provide a more methodical approach to word substitutions than random word substitutions for qualitative evaluations.

As seen in Table \ref{IMDb}, either replacing favorite with preferred or ``Kirsten Dunst'' with any of the listed actors/actresses suffices to change the classifier's prediction. Note that, ``Megan Fox's'' name appears in the same review in the previous sentence. Similar in Table  \ref{counterfac}, for counterfactual examples, it's sufficient to replace ``exciting'' with either an antonym (``boring'' or ``uninteresting'') or a synonym (``extraordinary'' or ``exceptional''). For easy examples however, despite trying to replace four or more high attribution words simultaneously with antonyms, the predicted sentiment did not change. Substitutions include ``good'' to ``bad'', ``unfunny'' to ``funny'', ``factually correct'' to ``factually incorrect''. Even though the passage included words like ``boredom'', a word that is generally associated with a negative movie review, the model did not assign it a high attribution score. Consequently, we did not try to substitute these words for testing robustness or fragility of word substitutions.

For our models, difficult examples have a mix of positive and negative words in a movie review. The models also struggled with examples of movies that selectively praise some attributes like acting (e.g., "Exceptional performance of the actors got me hooked to the movie from the beginning") while simultaneously use negative phrases (e.g., "however the editing was horrible"). Difficult examples also have high token attributions associated with irrelevant words like "nuclear," "get," and "an." Thus substituting one or two words in difficult examples change the predicted label of the classifier. Similarly, easier examples have clearly positive reviews (e.g., "Excellent direction, clever plot and gripping story"). Combining integrated gradients with high FIM examples can thus yield insights into the fragility of NLP models.

Several interesting observations emerge: CNN, Fasttext and LSTMs as a consequence of being trained on less amount of data is fragile to substitutions like Noun substitutions. BERT models on the other hand, because of their pretraining are robust to these sort of substitutions. High FIM BERT examples are more robust to meaningless changes, but single word substitutions still cause classifier prediction to change compared to low FIM examples.

\subsection{High $\lambda_{max}$ examples cause substantial drop in accuracy}
In order to study the effect of $\lambda_{max}$ on classification accuracy, we obtain examples with low FIM and high FIM $\lambda_{max}$ examples and plot classifier accuracy as a function of number $n$. As we see from Figure 3, low $\lambda_{max}$ examples retain accuracies upto 90 percent even after the test set has been replaced with perturbed examples. In the high $\lambda_{max}$ cases however, the accuracy drops belows 20 percent as we increase the number of examples.

We also investigated the effect of perturbation strength along largest eigenvector $e_{max}$ on classifier accuracy. As we see from Figure 3, part B, low $\lambda_{max}$ examples are significantly more robust to perturbations that high $\lambda_{max}$ examples, where even minimal perturbations of the weight of 0.1 can cause the classifier prediction to change.

\section{Discussion}
As evident from our experiments above, $\lambda_{max}$ is correlated with susceptibility to linguistic perturbations. Furthermore, $\lambda_{max}$ is directly correlated to the minimum perturbation needed to flip the classifier prediction. Finally, we show a stark difference in the response of low and high $\lambda_{max}$ examples, with the high $\lambda_{max}$ examples, perturbed examples being susceptible to meaningless perturbations (eg noun/synonym substitutions).

These experiments have several interesting ramifications: Firstly, it's risky to over rely on accuracy while studying the generalizability of NLP classifiers. Most of the classifiers we used in our experiments attain really high accuracy on the test set despite failing simple sanity checks (eg invariance to synonym substitutions). It's thus important to identify the examples which are most susceptible to perturbations and test their resilience. The fisher information provides us with a theoretically motivated framework to address this issue. By extracting the high $\lambda_{max}$ examples and perturbing them by using glove based synonym substitutions, we can construct a rolling test set for our NLP classifiers. Furthermore, by mining only the high $\lambda_{max}$ examples, FIM provides NLP practitioners with an interactive way to perturb and explore the flaws of their classifiers, something which would be extremely tedious to do on the entire dataset otherwise. While the focus of our work was not to generate adversarial examples, an alternate framing of this problem is to consider high $\lambda_{max}$ sentences to be more susceptible to adversarial attacks whereas low $\lambda_{max}$ sentences are generally robust to significant adversarial perturbations.

\begin{table}
\small
  \caption{Statistics of correlation between success probability and $\lambda_{max}$.}
  \label{counterfac}
  \centering
  \begin{tabular}{cccc}
  \hline
  \textbf{Dataset} & \textbf{Architecture} & \textbf{p-value} & \textbf{r-value}\\ \hline 
  YelpReviewPolarity & CNN &  1.32e-11 & 0.30 \\ 
  YelpReviewPolarity & FastText &  2.66e-18 & 0.38 \\ 
  YelpReviewPolarity & LSTM &  1.24e-18 & 0.38 \\ 
  \end{tabular}
\end{table}

\begin{table}
\small
  \caption{Statistics of correlation between success probability and $\lambda_{max}$.}
  \label{counterfac}
  \centering
  \begin{tabular}{cccc}
  \hline
  \textbf{Dataset} & \textbf{Architecture} & \textbf{p-value} & \textbf{r-value}\\ \hline 
  YelpReviewPolarity & CNN &  5.76e-42 & -0.411 \\ 
  YelpReviewPolarity & FastText &  7.66e-32 & -0.359 \\ 
  YelpReviewPolarity & LSTM &  7.75e-39 & -0.396 \\ 
  \end{tabular}
\end{table}

\section{Conclusion}
In this paper, we introduced a method to discover the highly fragile examples for an NLP classifier. our method discovered in several state of the art classifiers like BERT, CNN and LSTM. Furthermore, our experiments shed some light on the links between geometry of NLP classifiers and their linguistic and embedding space perturbability. Our method provides NLP practitioners with both an automated an interactive human in the loop framework to better understand their models.

There are several interesting extensions. Firstly, it would be interesting to decode the optimal perturbation vector $e_max$ associated with $\lambda_{max}$ as a sentence. We are developing several approaches based on finding token substitutions which maximize this dot product in order to address this. It will also be interesting to study the  link between purely geometrical properties like distance to the decision boundary and linguistic resilience. Finally, we are also interested in measuring the norm of token substitutions in embedding space to understand the relationship between the two.

\bibliography{uai2021-template}

\appendix

\section{Appendix}
\begin{figure*}[!ht]

\centering

\subfigure{\includegraphics[width=.33\textwidth]{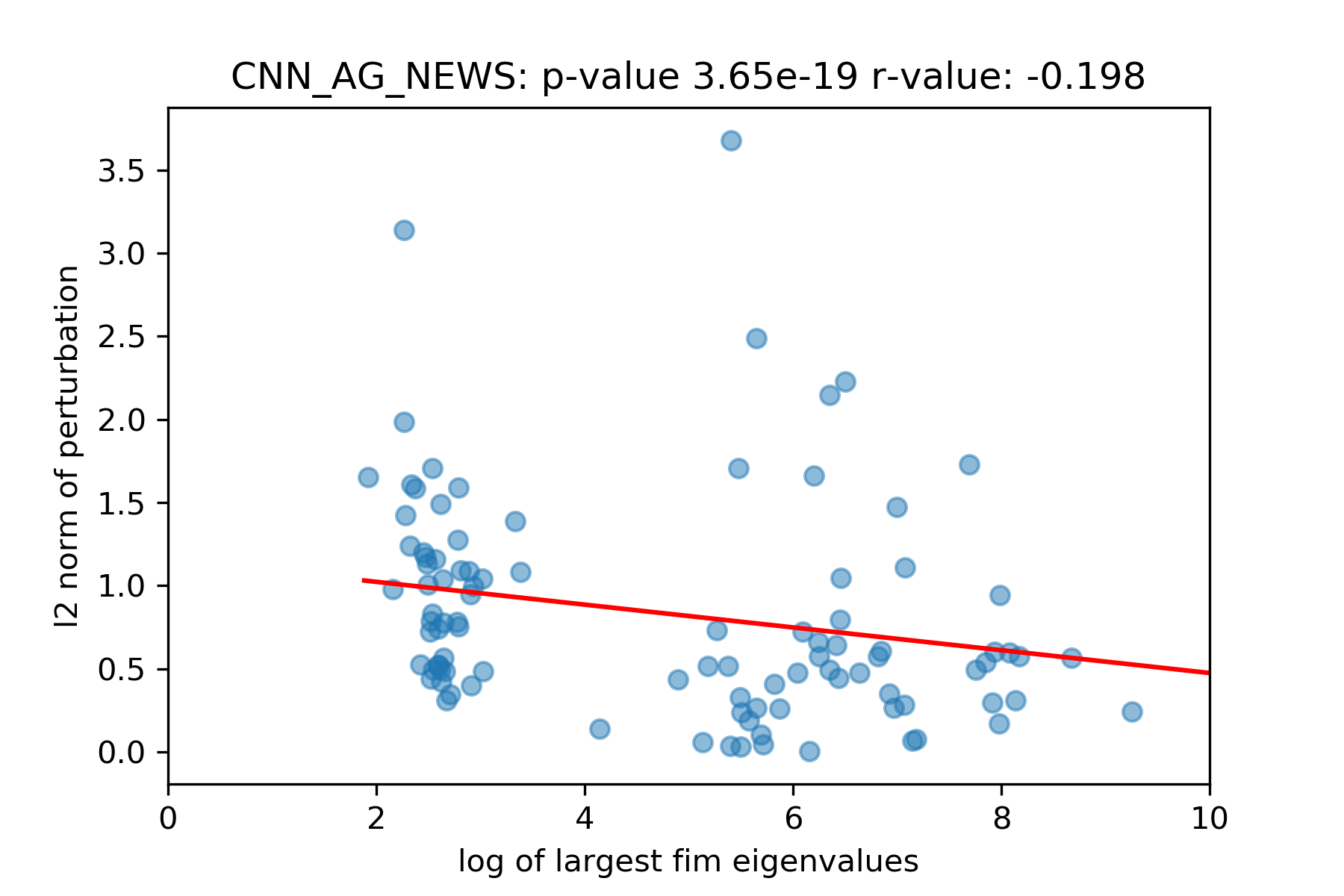}}
\subfigure{\includegraphics[width=.33\textwidth]{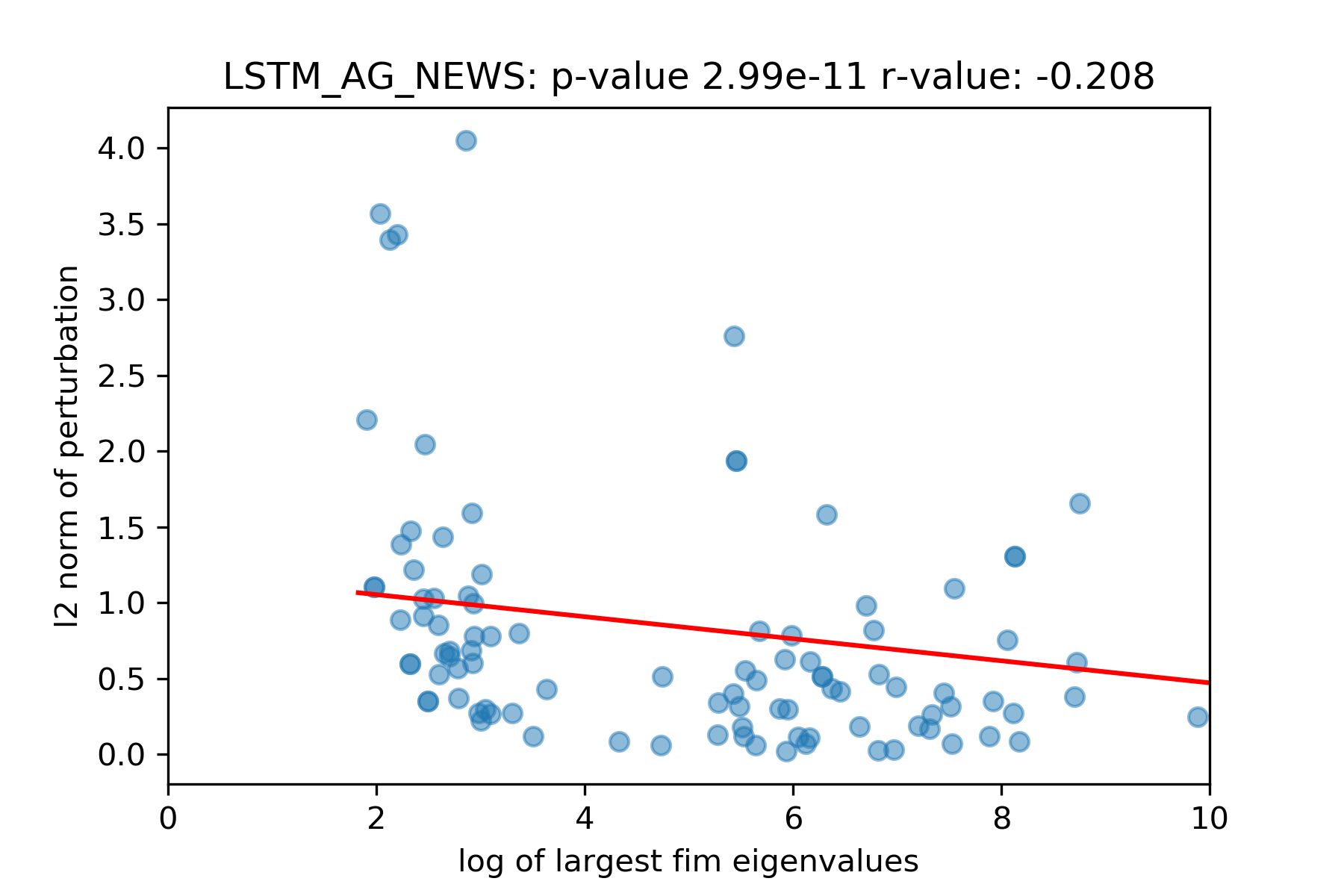}}
\subfigure{\includegraphics[width=.33\textwidth]{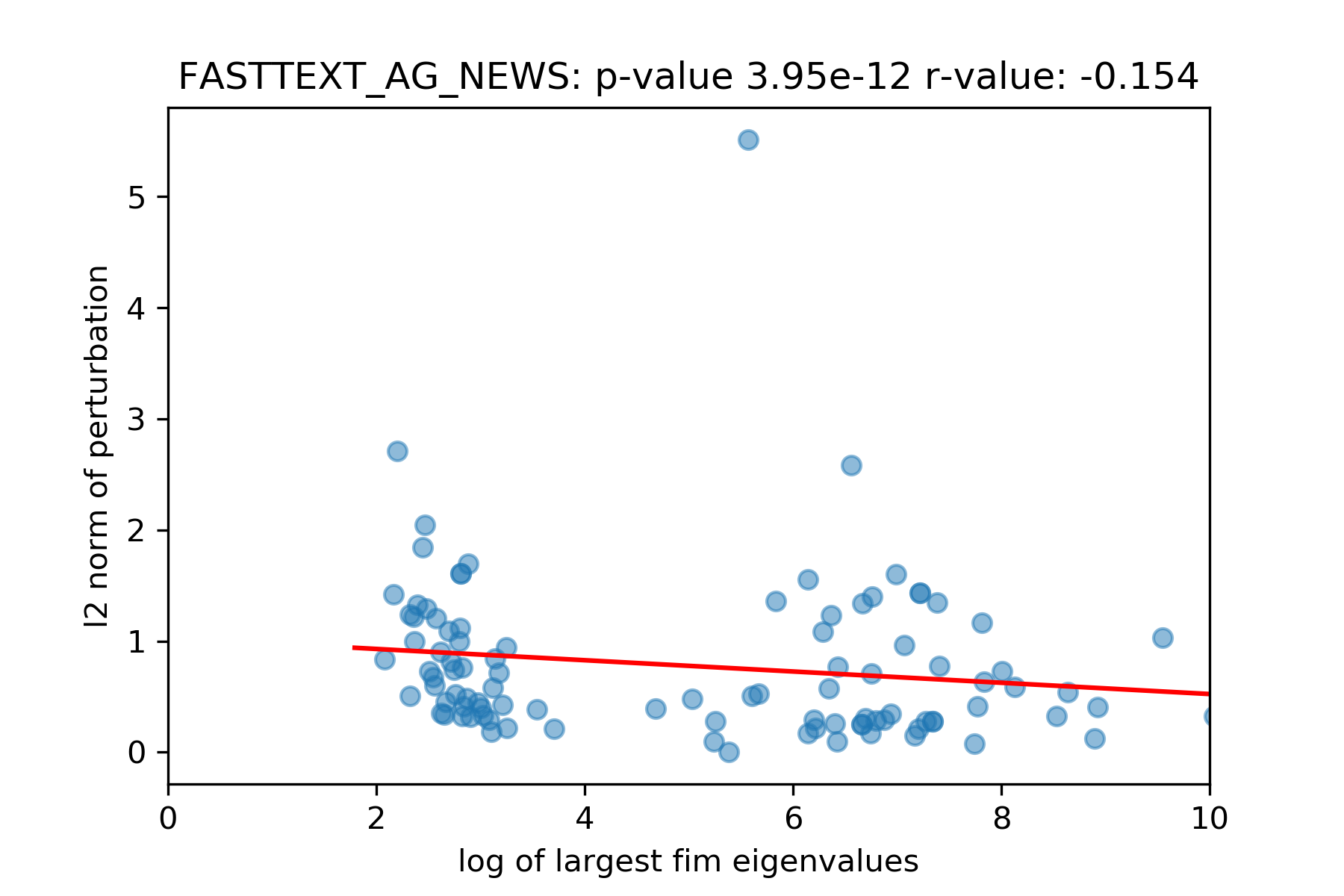}}
\label{counterfacHist}
\caption{Left figure: The difficult examples on the FIM test set are challenging for the classifiers with accuracy between 1-5\%. For the easy examples, (low FIM $\lambda_{max}$) the accuracy ranges between 60-100\%. Right figure: We plot the classifier accuracy with respect to the l2 norm of the perturbation in embedding space. For difficult examples, the classifier accuracy drops below 5\% at small perturbation strengths (l2 norm < 0.3) }
\end{figure*}

\begin{figure*}[ht]

\centering

\subfigure{\includegraphics[width=.45\textwidth]{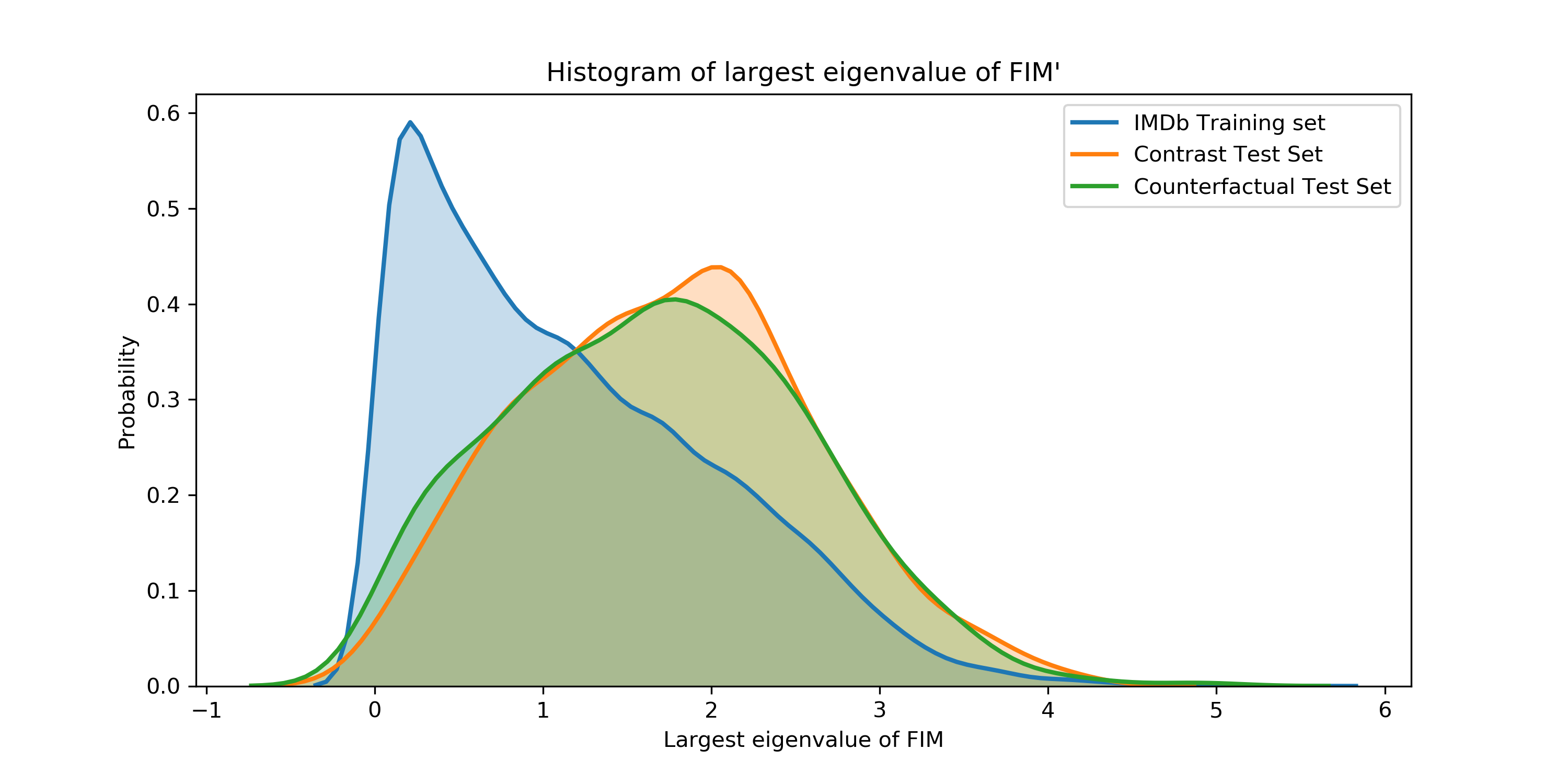}}
\subfigure{\includegraphics[width=.45\textwidth]{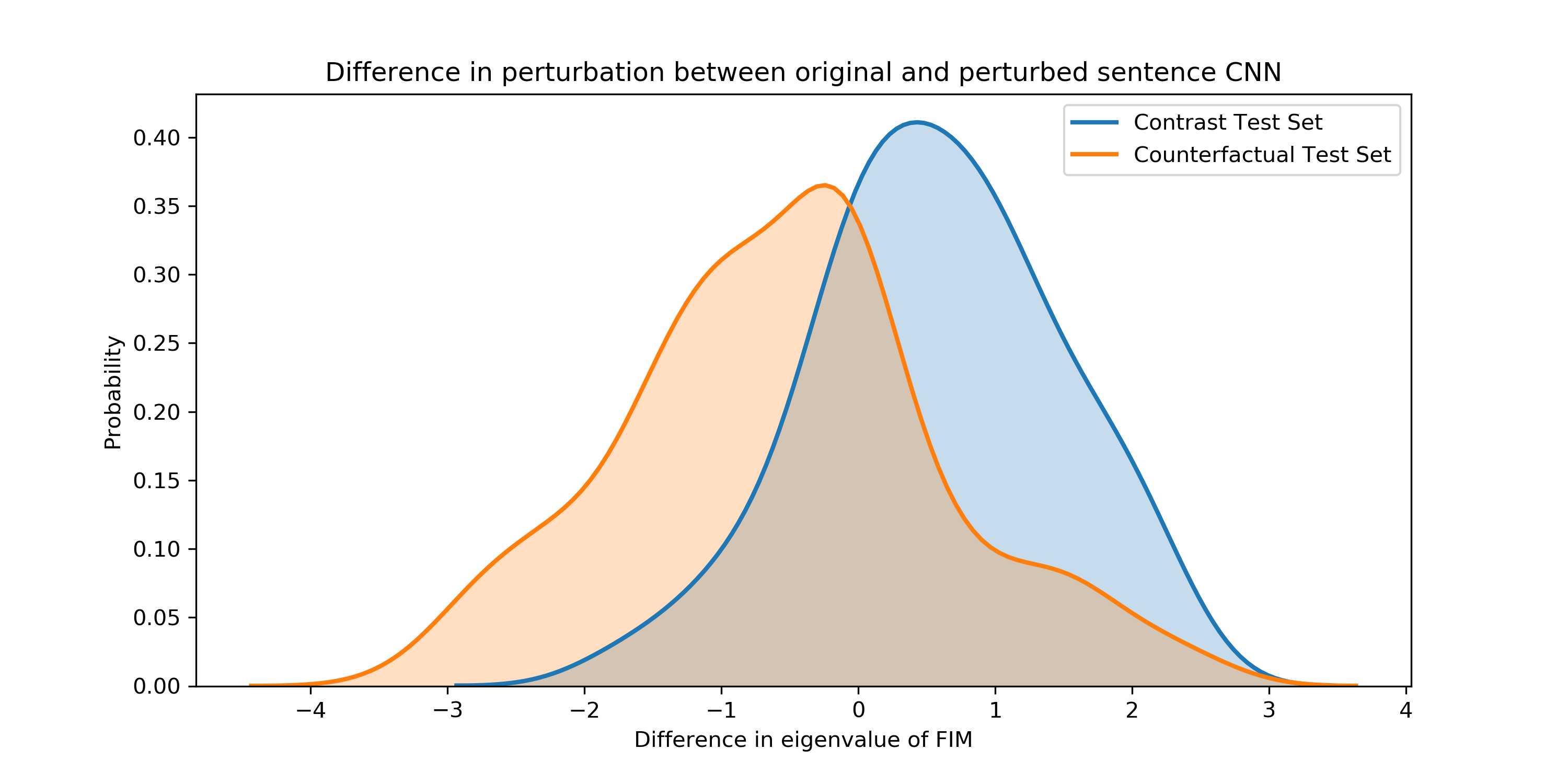}}

\label{counterfacHist}

\caption{a) Distribution of difference in largest eigenvalue of FIM of the original and the perturbed sentence in contrast set and counterfactual examples for BERT and CNN. With a mean near 0, these perturbations are not difficult for the model. Adhoc perturbations are thus not useful for evaluating model robustness. }
\end{figure*}

\end{document}